# THE DIGITAL RESTORATION OF DA VINCI'S SKETCHES

A. Sparavigna
Dipartimento di Fisica, Politecnico di Torino
Corso Duca Degli Abruzzi 24, Torino, Italy

**Abstract**
A sketch, found in one of Leonardo da Vinci's notebooks and covered by the written notes of this genius, has been recently restored. The restoration reveals a possible self-portrait of the artist, drawn when he was young. Here, we discuss the discovery of this self-portrait and the procedure used for restoration. Actually, this is a restoration performed on the digital image of the sketch, a procedure that can easily extended and applied to ancient documents for studies of art and palaeography.

**Introduction**
A sketch found in one of Leonardo da Vinci's notebooks could be a self-portrait. The drawing had been partially hidden by handwriting over it, and it remains unappreciated for 500 years before being discovered by Piero Angela, the most famous Italian scientific journalist [1,2]. The finding had been revealed on Italy's RAI television channel. After months of restoration work, an image was obtained showing the face of a young man. It is important to stress that the restoration was performed on a digital image of the document, not on the document itself. From the image, the written notes have been removed. Then the restoration procedure was completed by a slight reconstruction of image edges.
The portrait obtained after this procedure was aged using criminal investigation techniques and compared with the self-portrait of old Leonardo. After consultations with facial surgeons and police forensic experts, a general agreement arose on a strong similarity between the images.
This discovery of the self-portrait, which is of course very important by itself, is proposing a new application for image processing to the restoration of ancient documents. Let us then discuss deeply what happened with the Leonardo sketch and how the restoration was performed.

**The discovery**
On Saturday February 27, 2009, during a prime-time entertainment show of the RAI broadcaster on history and science presented by his son Alberto, Piero Angela showed how he discovered the self-portrait of Leonardo Da Vinci. This self-portrait is hidden in a page of the Da Vinci's Codex on the Flight of Birds. Now held at the Biblioteca Reale of Turin, the Codex on the Flight of Birds is a relatively small codex by Leonardo, dated approximately 1505. It comprises 18 folios and measures 21×15 centimetres. The codex begins with an examination of the flight behaviour of birds and proposes mechanisms for flying machines. Leonardo tested some of them, launching the flying machines from a hill near Florence, unfortunately, without success [3-5].
Angela said that when he was observing a copy of the manuscript written by Leonardo, he noticed that there was a drawing hidden between the words on the tenth page of the codex. First of all, the journalist noticed what looked like a nose underneath writing (see Fig.1). The fact that the drawing was made by a left-hand artist, as the directions of the sketching lines are showing, reinforced the necessity of further investigations. Enhancing the red-chalk sketch with the help a graphic artist, a portrait of a Renaissance man emerged. The restoration work on the image pixels revealed the face of a young man with long hair and a light beard. Angela realised that this could be a self-portrait of the young artist, and in fact, comparing the drawing with the Leonardo self portrait of c. 1512-15, the result is that the two men look like brothers.

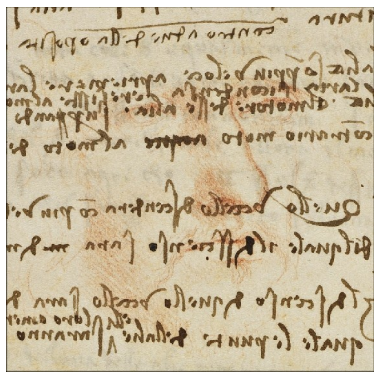

Fig.1 Piero Angela initially discovered what looked like a nose in a page of the Codex on the Flight of Birds.

The well-known researcher on Leonardo studies, Carlo Pedretti, agrees in considering the image as a self-portrait. Piero Angela explained the possible evolution of the artist's notebook as follows. Leonardo drew on the leaves, which constitute the eight central pages of the codex, between 1482 and 1489, when he was living at the court of Ludovico Sforza in Milan. Then these pages were 'recycled' by the artist to write his notes on the flight.

**Work on pixels**
In his notebooks, Leonardo employed a backwards form of writing that can be more easily read using the mirror images. He was left-handed and, probably, he just found easier and faster to write in this way. The notebooks were then written for his own use, with minimal organisation. As a consequence, it was quite natural for the artist to use sheets with previously sketched images for his notes on the flight. With the help of graphic artists, Angela obtained the restoration of one of these ghostly images, partially covered by the writing. After months of micro-pixel work, the graphic designers gradually cancelled the text by making it white instead of black, revealing the drawing beneath.

We can repeat this operation, to understand what happened during the manipulation of pixels. Each pixel can have red, green and blue tones ranging from 0 to 255. Let us evaluate histograms of the three colour tones, using the image in Fig.1. Histograms are shown in Fig.2. Each of them shows two peaks, one very high in the region of bright pixels and another little one in the region of dark pixels. Because the portrait is in red-chalk and the writing almost black, let us choose a threshold value for the red tone and set it between the two peaks of the red curve.

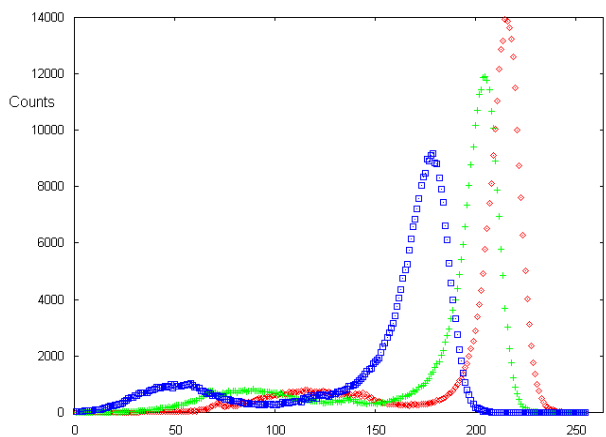

Fig.2. Histograms of red, green and blue tones of pixels in Fig.1. Each of the histograms shows two peaks, one very high in the bright region and a little one in the region of dark pixels.

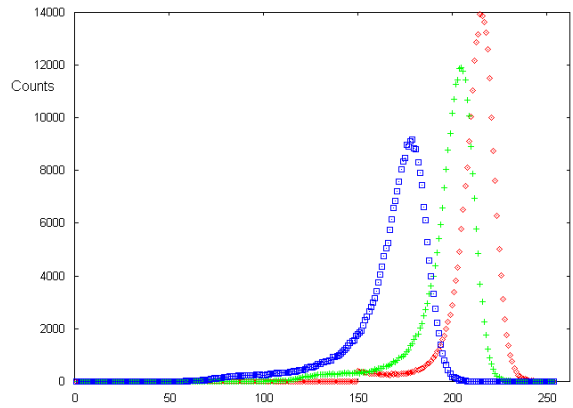

Fig.3. Histograms obtained after removing the written notes: the small peaks in the dark region disappear.

If the red tone has a value which is below the threshold value (for instance, 160), we replace the pixel with a white pixel. A new evaluation of histograms gives the result in Fig.3. We see from these new histograms, that the small peaks in the dark region have been removed: we can argue that these peaks were due to the pixels belonging to hand-written notes.

The new image on which we can work further is shown in Fig.4, on the right side of the original image: the hand-written text in now white. After removing the written letters, we can simply change the white pixels in pixels with the background colour tones. This procedure gives an image as that in the low part of Fig.4, on the left. The portrait gains more visibility in the grey image obtained using the blue channel only.

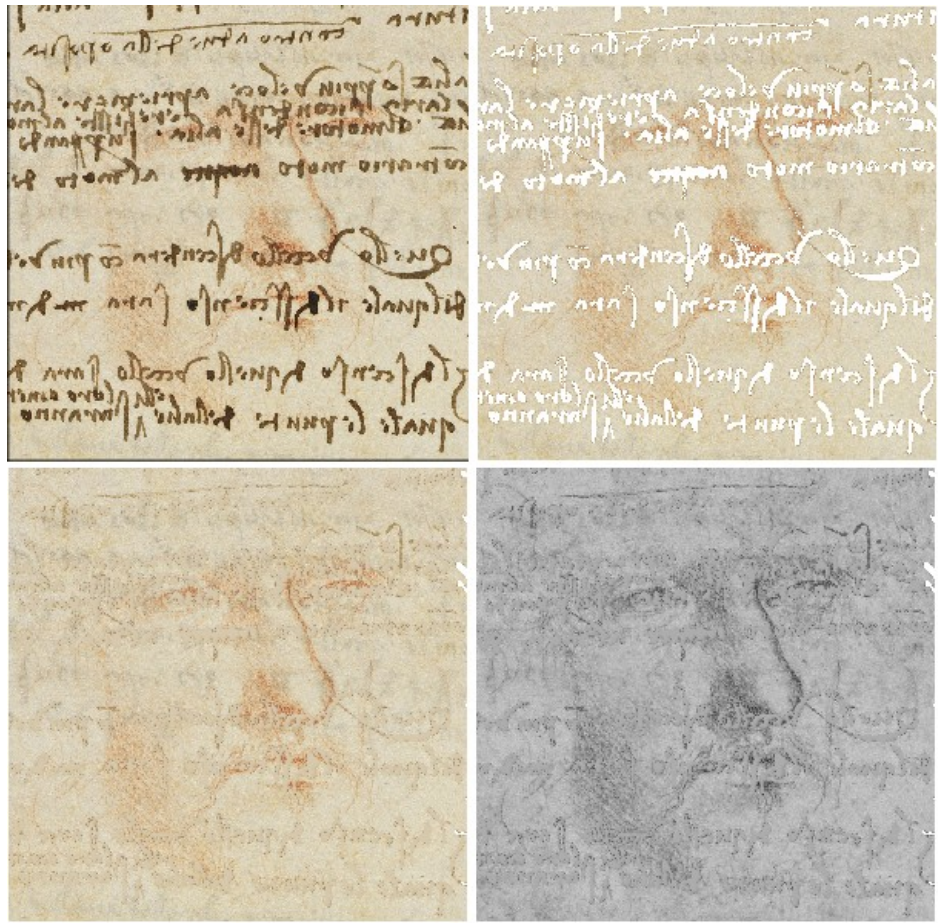

Fig.4. Removing the writing, we clearly see the image of a young man. According to Piero Angela, this is a self-portray of young Leonardo.

On the image obtained after removing the black pixels of hand-written text, we could apply one of the procedures of digital image restoration, which usually work on noisy and blurred images [6,7]. But we are not dealing with such problems and then we prefer to use a simple reconstruction of the white pixels, based only on the colour tones of nearest-neighbouring pixels. Let us consider then a white pixel: we replace it if three pixels in its nearest neighbour are not white. The new pixel has the colour tones given by the averaged values of these three pixels. The image on the left of Fig.5 was obtained with this reconstruction procedure, iterated until almost all the white pixels had been removed. On the right of Fig.5, we see the grey scale image obtained using only the blue tones. If we increase the value of the threshold (for instance instead of 160, we use a value of 190) some pixels of the drawing are also removed: this is the cost to have a better result (Fig.6).

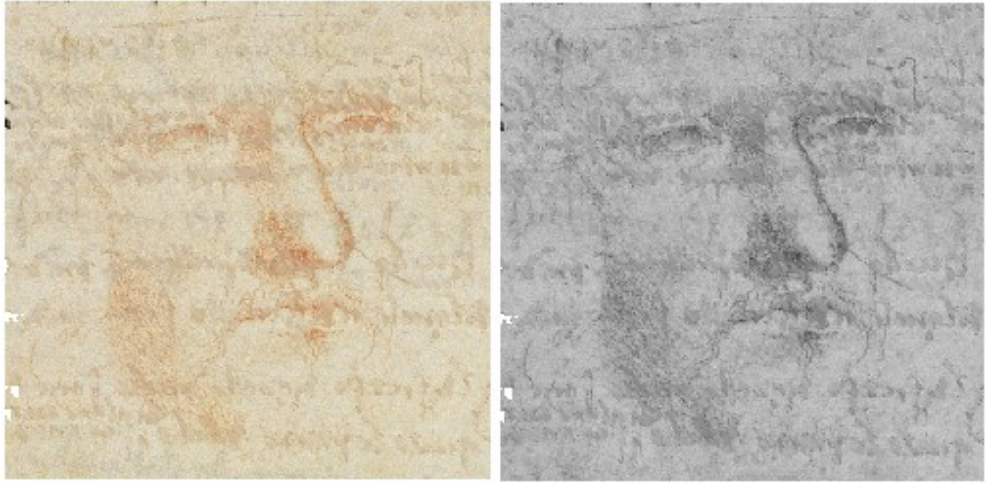

Fig.5 Result of interpolation with the nearest neighbouring pixels (threshold 160).

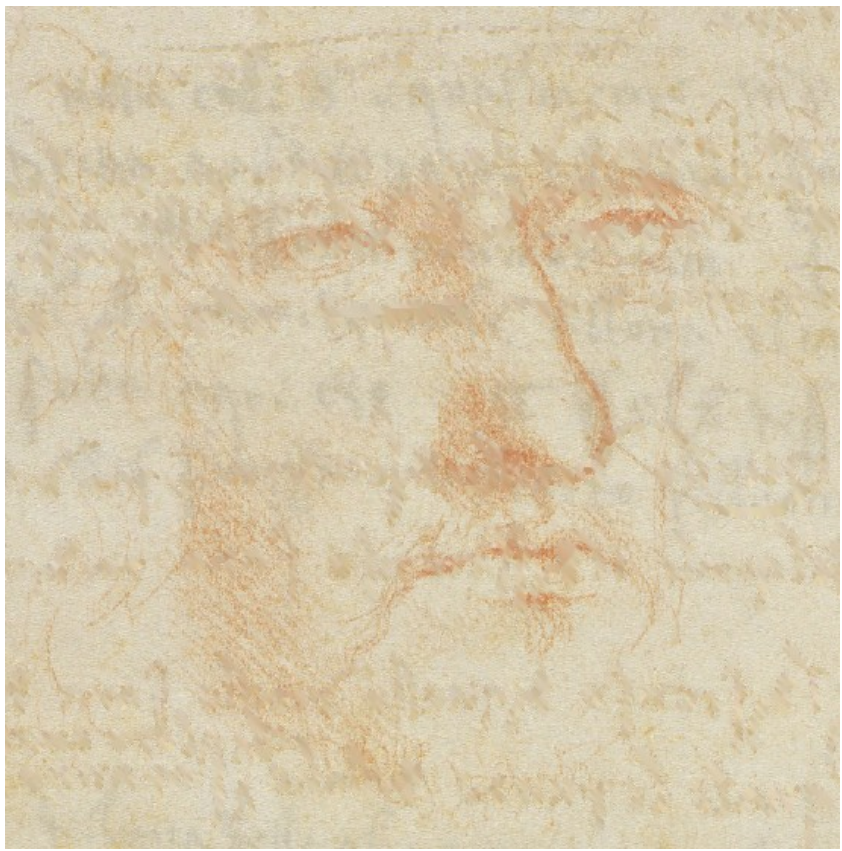

Fig.6 Result of interpolation with nearest neighbouring pixels, obtained with a higher threshold value.

The graphic artists engaged by Piero Angela obtained what we can consider a perfect result, after a great expenditure of micro-pixel work for several months (it is possible to admire their work at [8]). Here, we have proposed a simple approach based on interpolation with nearest neighbouring pixels, for the pleasure to repeat the discovery of Leonardo self-portrait. This procedure can be certainly improved, considering more refined interpolations, but this is beyond the aim of the paper. In fact, our aim is to stress the capabilities of digital processing in the restoration of ancient manuscripts. The digital procedure can be repeated for instance, on other Da Vinci's sketches, if the quality of image is good and the colour tones of ink different enough from the colour tones of the drawing.

**Conclusions**

We discussed the very recent discovery of a Leonardo self-portrait, which give us the picture of the young artist. This discovery is of course very important by itself. Because it was performed by means of a restoration of a high-resolution digital image, the discovery has another relevant value: it is showing the importance of digital restoration of old documents. This kind of restoration, which is not made on the document itself but on its digital image, can give excellent results when applied to images with proper characteristics (an example of restoration of an ancient code is shown in Fig.7). The discovery of Leonardo self-portrait can stimulate new applications for image processing or creation of new procedures, fundamental for studies of art and palaeography.

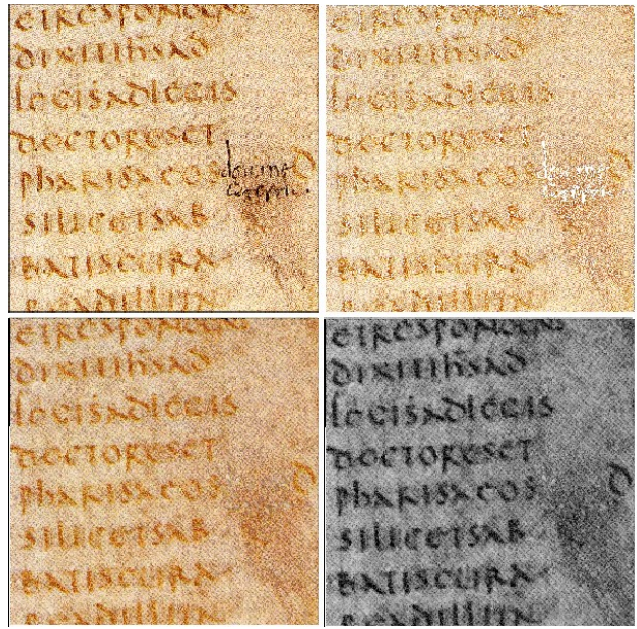

Fig.7. A small part in a page of the Code Eusebii, at the Biblioteca Capitolare, Vercelli, before and after restoration [9].